\documentclass[10pt, a4paper]{article}

\usepackage[final]{lrec2026}

\usepackage{latexsym}
\usepackage{amsthm}
\usepackage{booktabs}
\usepackage{enumitem}
\usepackage{graphicx}
\usepackage{color}
\usepackage[utf8]{inputenc}
\usepackage[T1]{fontenc}
\usepackage{tocbasic}
\usepackage{multicol}
\usepackage{multirow}
\usepackage{makecell}
\usepackage{array}
\usepackage{subcaption}
\usepackage{bm}
\usepackage[dvipsnames,table]{xcolor}
\usepackage{tikz}
\usetikzlibrary{arrows.meta,positioning,calc}
\usepackage{longtable}
\usepackage{adjustbox}
\usepackage{ltablex}
\usepackage{placeins}
\usepackage{hyperref}
\usepackage{arydshln}
\definecolor{NavyBlue}{RGB}{52, 84, 209}
\definecolor{DarkAmber}{RGB}{255, 170, 0}
\definecolor{ModernPurple}{RGB}{140, 90, 200}

\title{Toward Generalized Cross-Lingual Hateful Language Detection with Web-Scale Data and Ensemble LLM Annotations}

\name{Dang H.\ Dang, Jelena Mitrovi\'{c}, Michael Granitzer}

\address{University of Passau, Chair of Data Science \\
         \{hai-dang.dang, jelena.mitrovic, michael.granitzer\}@uni-passau.de}

\abstract{
We study whether large-scale unlabelled web data and LLM-based synthetic annotations can improve multilingual hate speech detection.
Starting from texts crawled via OpenWebSearch.eu~(OWS) in four languages (English, German, Spanish, Vietnamese), we pursue two complementary strategies.
First, we apply \emph{continued pre-training} to BERT models by continuing masked language modelling on unlabelled OWS texts before supervised fine-tuning, and show that this yields an average macro-F1 gain of approximately 3\% over standard baselines across sixteen benchmarks, with stronger gains in low-resource settings.
Second, we use four open-source LLMs (\texttt{Mistral-7B}, \texttt{Llama3.1-8B}, \texttt{Gemma2-9B}, \texttt{Qwen2.5-14B}) to produce synthetic annotations through three ensemble strategies: mean averaging, majority voting, and a LightGBM meta-learner.
The LightGBM ensemble consistently outperforms the other strategies.
Fine-tuning on these synthetic labels substantially benefits a small model (\texttt{Llama3.2-1B}: +11\% pooled F1), but provides only a modest gain for the larger \texttt{Qwen2.5-14B} (+0.6\%).
Our results indicate that the combination of web-scale unlabelled data and LLM-ensemble annotations is the most valuable for smaller models and low-resource
languages.
\\ \newline }

\begin{document}

\maketitleabstract

\section{Introduction}

A central bottleneck in building robust detectors for hateful and offensive language is the scarcity of high-quality labelled training data~\cite{vidgen2020directions, fortuna2018survey}.
While raw web text can be collected at scale, annotating it remains costly~\cite{ross2017measuring} and human annotators inevitably introduce subjective biases~\cite{caselli2021hatebertretrainingbertabusive}.
Recent large language models (LLMs) have demonstrated strong performance on hate-speech benchmarks~\cite{guo2024investigationlargelanguagemodels} and have therefore been explored as automated annotators~\cite{hartvigsen2022toxigenlargescalemachinegenerateddataset}.
However, existing work in this space remains narrow in scope, typically covering a single language and lacking rigorous comparison with models trained on human labels.

Large web crawls such as OpenWebSearch.eu~(OWS)~\cite{openwebsearch} and OpenWebIndex~(OWI)~\cite{10.1007/978-3-031-56069-9_10} make billions of multilingual pages available, yet how to best leverage them for hate-speech detection is an open question.
We address this gap along two axes:
\begin{enumerate}[leftmargin=*,nosep]
    \item We investigate whether \emph{domain-adaptive continued pre-training} on large unlabelled OWS corpora improves the downstream performance of BERT-family models.
    \item We study whether \emph{ensemble-based LLM annotation} can replace or supplement human labelling for multilingual hate-speech detection.
\end{enumerate}

\paragraph{Terminology.}
Throughout this paper, \textbf{continued pre-training} denotes an additional masked-language-modelling adaptation step applied to a general-purpose pre-trained BERT using domain-relevant but \emph{unlabelled} OWS texts.
This precedes the subsequent supervised \textbf{fine-tuning} on hate-speech labelled data.
The term follows the domain-adaptive pre-training (DAPT) paradigm of \citet{gururangan2020don}.
Figure~\ref{fig:pipeline} summarises the end-to-end pipeline.

Concretely, this work is guided by two research questions:
\begin{itemize}[leftmargin=*,nosep]
    \item \textbf{RQ1:} To what extent can large-scale unlabelled web data improve multilingual hate-speech detection via continued pre-training of BERT?
    \item \textbf{RQ2:} How effective are LLM ensemble strategies for synthetic annotation in improving detection performance and cross-lingual generalisation?
\end{itemize}

\noindent
Code, fine-tuned models, OWS data, and labelled splits are available at \url{https://github.com/HaiDangDang/HateOWS}.\footnote{Repository includes all training scripts, annotation prompts, and evaluation code.}

\section{Related Work}

\paragraph{Domain-specific continued pre-training for hate speech.}
\citet{gururangan2020don} established that \emph{domain-adaptive pre-training} (DAPT), continuing masked language modelling on in-domain text before fine-tuning, consistently improves downstream performance.
HateBERT~\cite{caselli2021hatebertretrainingbertabusive} applied DAPT to abusive language using 1.5M English Reddit posts, while \citet{belay2025afroxlmrsocial} extended it multilingually with AfroXLMR-Social across 19~African languages (F1 gains of 1--30\%).
None of these works use web-crawled data at OWS scale or cover our four-language setting.

\paragraph{Cross-lingual hate speech detection.}
\citet{dmonte2024generalizedoffensivelanguageidentification} showed via the GenOffense benchmark that single-dataset models generalise poorly, though their study was English-only.
\citet{usman2025multilingualhatespeech} explored translation-based LLM approaches for multilingual detection, highlighting the difficulty of cross-lingual transfer.
We extend these evaluations to four languages and sixteen benchmarks.

\paragraph{Ensemble approaches.}
Kashif et al.~\cite{kashif2023lexicalsquadmultimodalhatespeech} and Daouadi et al.~\cite{daouadi2024ensemblepretrainedlanguagemodels} built multimodal and Arabic BERT ensembles respectively for downstream classification.
Our work instead applies the ensemble paradigm to the \emph{annotation} phase, combining token-probability outputs from four LLMs.

\paragraph{LLM-based annotation and synthetic data.}
\citet{zhu2023chatgptlabels} found ChatGPT achieves 60.9\% average accuracy relabelling social-computing datasets, while \citet{giorgi2025humanllmbiases} showed that LLM annotation biases differ substantially from human biases, and \citet{piot2025llmreliability} demonstrated that LLMs can reliably reproduce model-ranking patterns as scalable proxy evaluators.
Our work \emph{ensembles} four open-source LLMs for annotation and compares three aggregation strategies, measuring how synthetic labels benefit models of varying capacity.

\section{Methodology and Setup}

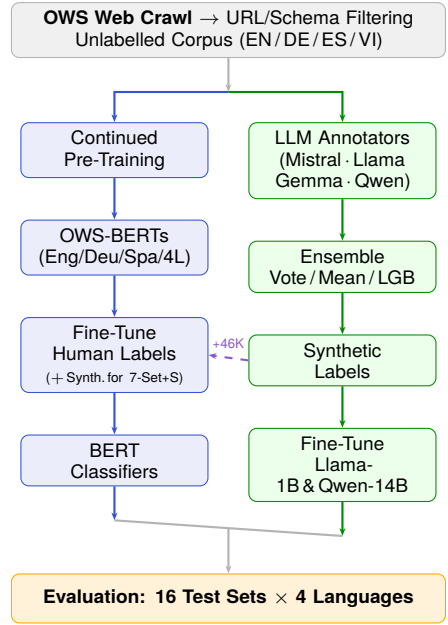
\begin{figure}[ht]
\centering
\begin{tikzpicture}[
    node distance = 0.55cm,
    rect/.style  = {rectangle, rounded corners=3pt, align=center,
                    font=\scriptsize, minimum height=0.65cm, inner sep=3pt},
    ows/.style   = {rect, text width=5.5cm,
                    draw=gray!60, fill=gray!12},
    bert/.style  = {rect, text width=2.25cm,
                    draw=NavyBlue, fill=NavyBlue!10},
    llm/.style   = {rect, text width=2.25cm,
                    draw=green!55!black, fill=green!8},
    eval/.style  = {rect, text width=5.5cm, font=\scriptsize\bfseries,
                    draw=DarkAmber, fill=DarkAmber!15},
    ga/.style    = {-{Stealth[length=3pt]}, thick, gray!60},
    ba/.style    = {-{Stealth[length=3pt]}, thick, NavyBlue},
    la/.style    = {-{Stealth[length=3pt]}, thick, green!55!black},
]

\node[ows] (corpus)
  {\textbf{OWS Web Crawl} $\to$ URL/Schema Filtering\\
   Unlabelled Corpus (EN\,/\,DE\,/\,ES\,/\,VI)};

\node[bert, below=0.85cm of corpus, xshift=-1.5cm] (pretrain)
  {Continued\\Pre-Training};
\node[bert, below=of pretrain] (owsbert)
  {OWS-BERTs\\(Eng/Deu/Spa/4L)};
\node[bert, below=of owsbert] (humanlbl)
  {Fine-Tune\\Human Labels\\{\tiny($+$\,Synth.\,for 7-Set+S)}};
\node[bert, below=of humanlbl] (bertout)
  {BERT\\Classifiers};

\node[llm, below=0.85cm of corpus, xshift=+1.5cm] (llmanno)
  {LLM Annotators\\(Mistral\,\textperiodcentered\,Llama\\Gemma\,\textperiodcentered\,Qwen)};
\node[llm, below=of llmanno] (ens)
  {Ensemble\\Vote\,/\,Mean\,/\,LGB};
\node[llm, below=of ens] (synthlbl)
  {Synthetic\\Labels};
\node[llm, below=of synthlbl] (llmft)
  {Fine-Tune\\Llama-1B\,\&\,Qwen-14B};

\coordinate (lend) at ($(bertout.south)+(0,-0.45)$);
\coordinate (rend) at ($(llmft.south)+(0,-0.45)$);
\coordinate (mend) at ($(lend)!0.5!(rend)$);
\node[eval, below=0.6cm of mend] (eval)
  {Evaluation: 16 Test Sets $\times$ 4 Languages};

\draw[ga] (corpus.south) -- ++(0,-0.45) coordinate (split);
\draw[ba] (split) -| (pretrain.north);
\draw[la] (split) -| (llmanno.north);
\draw[ba] (pretrain) -- (owsbert);
\draw[ba] (owsbert) -- (humanlbl);
\draw[ba] (humanlbl) -- (bertout);
\draw[la] (llmanno) -- (ens);
\draw[la] (ens) -- (synthlbl);
\draw[la] (synthlbl) -- (llmft);
\draw[ModernPurple, thick, dashed, -{Stealth[length=3pt]}]
  (synthlbl.west) -- (humanlbl.east)
  node[midway, above, font=\tiny, text=ModernPurple] {+46K};
\draw[ba] (bertout.south) -- (lend);
\draw[la] (llmft.south)  -- (rend);
\draw[gray!60, thick] (lend) -- (rend);
\draw[ga] (mend) -- (eval.north);

\end{tikzpicture}
\caption{End-to-end pipeline.
  \textcolor{NavyBlue}{\textbf{Left (BERT track)}}: OWS texts are used for domain-adaptive continued pre-training; models are then fine-tuned on human labels.
  \textcolor{green!55!black}{\textbf{Right (LLM track)}}: four LLMs generate token-probability annotations; three ensemble strategies produce synthetic labels used to fine-tune smaller LLMs.
  The \textcolor{ModernPurple}{dashed arrow} shows the cross-track path: ${\approx}$46K \texttt{Qwen2.5-14B}-annotated texts augment the BERT training data (7-Set\,+\,Synth.).
  Both tracks are evaluated on the same sixteen test sets.
  LGB\,=\,LightGBM.}
\label{fig:pipeline}
\end{figure}

Figure~\ref{fig:pipeline} provides a high-level overview of the pipeline.
The following subsections detail each component.

\subsection{OpenWebSearch.eu Data Collection}

We used OWS~\cite{openwebsearch} and OWI~\cite{10.1007/978-3-031-56069-9_10} to collect large-scale web texts in English, German, Spanish, and Vietnamese.
To increase the proportion of conversational and user-generated content, we filtered the OWS index to retain only URLs whose path contained at least one of the keywords \textit{thread}, \textit{forum}, \textit{reply}, \textit{post}, \textit{status update}, or \textit{quote}.
We additionally required retrieved pages to conform to one of ten conversational schema.org types (e.g.\ \texttt{DiscussionForumPosting}, \texttt{SocialMediaPosting}, \texttt{Comment}; full list in Appendix~\ref{app:ows-schema}), to comply with structured human discourse.
\textbf{No hate-specific keyword filtering was applied}: the unlabelled texts serve purely for domain-adaptive continued pre-training, not as a labelled hate-speech corpus.

In total, we collected approximately 6M English, 3M German, 3M Spanish, and 500K Vietnamese texts.
We sub-sampled differently per task: ${\sim}3$M mixed-language texts for BERT continued pre-training and ${\sim}2$M texts for LLM-based annotation.


\subsection{Human-Labelled Datasets}

\begin{table}[ht]
    \centering
    \resizebox{\linewidth}{!}{
        \begin{tabular}{l|cc|cc|c|c}
        \hline
        \textbf{Dataset} & \multicolumn{2}{c|}{\textbf{Train}} &
        \multicolumn{2}{c|}{\textbf{Test}} \\
        \cmidrule(lr){2-3} \cmidrule(lr){4-5}
        &  Inst\# & \%\,Hate & Inst\# &  \%\,Hate & Lang & Ref \\
        \hline
        \textcolor{RoyalBlue!80!black}{\textbf{HateXplain}} & 15,299 & 0.59 & 3,846 & 0.59 & eng & \citep{mathew2022hatexplainbenchmarkdatasetexplainable} \\
        \textcolor{RoyalBlue!80!black}{\textbf{Sexism}} & 10,904 & 0.13 & 2,632 & 0.13 & eng & \citep{samory2021callsexistbutrevisiting} \\
        \textcolor{RoyalBlue!80!black}{\textbf{Covid}} & 1,282 & 0.19 & 971 & 0.20 & eng & \citep{covid19dataset} \\
        \textcolor{RoyalBlue!80!black}{\textbf{US\_election}} & 1,283 & 0.12 & 1,117 & 0.13 & eng & \citeyearpar{grimminger-klinger-2021-hate} \\
        \textcolor{RoyalBlue!80!black}{HateEval-eng} & 9,000 & 0.42 & 1,000 & 0.43 & eng & \citep{basile-etal-2019-semeval} \\
        \textcolor{RoyalBlue!80!black}{AbusEval} & 13,240 & 0.21 & 860 & 0.21 & eng & \citep{caselli-etal-2020-feel} \\
        \textcolor{RoyalBlue!80!black}{AHSD} & 21,783 & 0.83 & 3,000 & 0.83 & eng & \citep{davidson2017automatedhatespeechdetection} \\
        \textcolor{RoyalBlue!80!black}{Eng-All} & 72,791 & 0.49 & 13,426 & 0.19 & eng &  \\
        \midrule
        \textcolor{SeaGreen!80!black}{\textbf{GermEval21}} & 2,071 & 0.33 & 2,085 & 0.37 & deu & \citep{risch-etal-2021-overview} \\
        \textcolor{SeaGreen!80!black}{\textbf{GermEval19}} & 9,698 & 0.33 & 2,507 & 0.34 & deu & \citep{423423523} \\
        \textcolor{SeaGreen!80!black}{GermEval18} & 5,009 & 0.34 & 3,532 & 0.34 & deu & \citep{DATA/0B5VML_2019} \\
        \textcolor{SeaGreen!80!black}{HASOC} & 2,373 & 0.28 & 526 & 0.25 & deu & \citep{10.1145/3368567.3368584} \\
        \textcolor{SeaGreen!80!black}{Gahd} & 8,797 & 0.42 & 2,198 & 0.43 & deu & \citep{goldzycher-etal-2024-improving} \\
        \textcolor{SeaGreen!80!black}{Deu-All} & 27,948 & 0.36 & 10,848 & 0.09 & deu &  \\
        \midrule
        \textcolor{BurntOrange!90!black}{\textbf{ViHSD}} & 8,061 & 0.17 & 2,672 & 0.18 & vie & \citep{Luu_2021} \\
        \midrule
        \textcolor{Orchid!80!black}{Haternet} & 4,794 & 0.26 & 1,205 & 0.25 & spa & \citep{basile-etal-2019-semeval} \\
        \textcolor{Orchid!80!black}{HateEval-spa} & 5,309 & 0.41 & 1,286 & 0.43 & spa & \citeyearpar{s19214654} \\
        \textcolor{Orchid!80!black}{Chileno} & 7,572 & 0.06 & 1,928 & 0.06 & spa & \citeyearpar{arango-monnar-etal-2022-resources} \\
        \textcolor{Orchid!80!black}{Spa-All} & 17,675 & 0.22 & 4,419 & 0.02 & spa &  \\
        \hline
        \end{tabular}
    }
    \caption{Sixteen human-labelled hate-speech datasets across four languages.
    Inst\# = number of instances; \%\,Hate = fraction labelled hateful.
    Datasets in \textbf{bold} form the \emph{7-Set} training subset.}
    \label{tab:dataset_comparison}
\end{table}

We used sixteen publicly available hate-speech datasets (Table~\ref{tab:dataset_comparison}) spanning four languages.
The datasets vary in annotation guidelines, topic coverage, and label granularity.

\paragraph{Label mapping.}
To enable unified training and evaluation, we mapped all annotations to a binary \textbf{Hate}/\textbf{Neutral} scheme.
Three datasets (\textit{HateXplain}, \textit{AHSD}, \textit{ViHSD}) originally distinguish \textit{Hate} from \textit{Offensive}; both were merged into \textbf{Hate}.
The \textit{AbusEval} labels \textit{Implicit Abusive} and \textit{Explicit Abusive} were similarly collapsed.
The remaining twelve datasets already use binary labels.

We acknowledge that merging qualitatively different categories (hate, offensiveness, and abuse) into a single positive class is a simplification.
It may inflate apparent recall on datasets whose original positive class was broader (e.g., \textit{AbusEval}).
Readers should keep this caveat in mind when interpreting cross-dataset comparisons.

\paragraph{Training configurations.}
We defined five configurations: (i)~\emph{7-Set}---seven datasets (bold in Table~\ref{tab:dataset_comparison}), excluding all Spanish data, used for \texttt{Llama3.2-1B}~\cite{grattafiori2024llama3herdmodels} and \texttt{Qwen2.5-14B}~\cite{qwen2025qwen25technicalreport} fine-tuning; (ii)~\emph{Eng}, (iii)~\emph{Deu}, (iv)~\emph{Spa}---language-specific subsets for BERT experiments; and (v)~\emph{16-Mix}---all sixteen datasets combined.
For evaluation, we used the test portions of all sixteen datasets.

\subsection{BERT Continued Pre-Training and Fine-Tuning}

We continually pre-trained four BERT variants on OWS texts:

\begin{itemize}[leftmargin=*,nosep]
    \item \texttt{OwsSpa}: 1.1M Spanish texts (${\approx}$110M tokens).
    \item \texttt{OwsDeu}: 650K German texts (${\approx}$52M tokens).
    \item \texttt{OwsEng}: 1.5M English texts (${\approx}$119M tokens).
    \item \texttt{Ows4L}: 2.78M multilingual texts---43\% English, 32\% German, 17\% Spanish, 6.7\% Vietnamese (${\approx}$136M tokens).
\end{itemize}

Each model was then \emph{supervised-fine-tuned} on its corresponding training configuration.
\texttt{Ows4L} was evaluated under three settings: \emph{7-Set}, \emph{16-Mix}, and \emph{7-Set + Synthetic} (adding ${\approx}$46K \texttt{Qwen2.5-14B}-annotated OWS examples; see Section~\ref{sec:llm-annotation}).

\subsection{LLM Ensemble Annotation}\label{sec:llm-annotation}

\subsubsection{Annotation Method}\label{annotation_method}

We collected token-level output probabilities from four instruction-tuned, 4-bit quantised LLMs via the Unsloth framework~\cite{unslothai}: \texttt{Mistral-7B}~\cite{jiang2023mistral}, \texttt{Llama3.1-8B}, \texttt{Gemma2-9B}~\cite{gemmateam2024gemma2improvingopen}, and \texttt{Qwen2.5-14B}.
A simple zero-shot prompt was used to classify each OWS text as \textit{Hate} or \textit{Neutral}.
From the raw probabilities, three ensemble strategies determine the final label:

\begin{enumerate}[leftmargin=*,nosep]
    \item \textbf{Majority Voting (\texttt{Vote}):} Each model's probability is thresholded to a hard label; a text is marked \textit{Hate} if at least two of the four models vote for it.
    \item \textbf{Mean Averaging (\texttt{Mean}):} For each class, the average probability across all models is computed; the class with the higher mean is assigned.
    \item \textbf{LightGBM Meta-Learner (\texttt{LGB}):} A LightGBM classifier~\cite{NIPS2017_6449f44a} is trained on the eight-dimensional probability vectors (two classes $\times$ four models) using the seven human-labelled training sets as supervision.
    Unlike \texttt{Vote} and \texttt{Mean}, which treat all models equally, \texttt{LGB} learns to weight each annotator differentially based on its reliability, and discovers confidence thresholds that align with human judgements.
    This discriminative calibration explains its consistent advantage (Section~\ref{sec:rq2}).
\end{enumerate}

\paragraph{Label imbalance in synthetic data.}
A notable characteristic of the resulting annotations is severe class imbalance: \texttt{Neutral} comprises over 97\% of annotated OWS texts across all three ensemble methods.
This reflects the web's base rate (most discussion-forum content is not hateful) and directly constrains the minority-class detection ability of models trained on these sets.

\subsubsection{LoRA Fine-Tuning of LLMs}

We used Low-Rank Adaptation (LoRA)~\cite{hu2021loralowrankadaptationlarge} to fine-tune \texttt{Llama3.2-1B} and \texttt{Qwen2.5-14B}.
Human-labeled training used the \emph{7-Set} (excluding Spanish).
The synthetic subset contains \textbf{240,647} texts: 125,617 German, 108,375 English, and 6,655 Vietnamese.
After ensemble annotation, \texttt{Vote} assigned 4,717 texts as hate, \texttt{Mean} 3,994, and \texttt{LGB} 3,122---confirming the pronounced class imbalance.

\subsubsection{Computational Resources}\label{sec:Resource}

All experiments ran on a single NVIDIA A6000 (48\,GB).
Table~\ref{tab:compute} summarises the cost per task.

\begin{table}[ht]
\centering
\footnotesize
\setlength{\tabcolsep}{4pt}
\begin{tabular}{@{}lrrr@{}}
\toprule
\textbf{Task / Model} & \textbf{Time} & \textbf{VRAM} & \textbf{Params} \\
\midrule
\multicolumn{4}{@{}l}{\textit{Annotation (240K, batch 128)}} \\
\quad \texttt{Mistral-7B} & 4\,h & 13\,GB & -- \\
\quad \texttt{Llama3.1-8B} & 4\,h\,45\,m & 23\,GB & -- \\
\quad \texttt{Gemma2-9B} & 6\,h\,30\,m & 43\,GB & -- \\
\quad \texttt{Qwen2.5-14B} & 7\,h & 42\,GB & -- \\
\midrule
\multicolumn{4}{@{}l}{\textit{LoRA Fine-Tuning}} \\
\quad \texttt{Llama3.2-1B} & ${<}$2\,h & ${<}$5\,GB & 5.6M\,(0.45\%) \\
\quad \texttt{Qwen2.5-14B} & ${\approx}$12\,h & ${<}$15\,GB & 34.4M\,(0.23\%) \\
\bottomrule
\end{tabular}
\caption{Cost per task on a single NVIDIA A6000 (48\,GB).}
\label{tab:compute}
\end{table}

\section{Results}

\subsection{RQ1: OWS Continued Pre-Training with BERT}

Table~\ref{tab:BertCompare} reports macro-F1 across all sixteen test sets.
All four OWS-continually-pre-trained models outperform both BERT and HateBERT on every per-language average and on the overall 16-set average under multilingual training (7-Set Mix, 7-Set + Synth., 16-Mix).
Monolingual OWS models additionally beat both baselines on their target-language test sets.

\subsubsection{Single-Language Continued Pre-Training}

\paragraph{In-language performance.}
When fine-tuned on language-matched data, all three monolingual OWS models outperform both BERT and HateBERT on their target language (shaded cells in Table~\ref{tab:BertCompare}).
\texttt{OwsEng} reaches an English average of 86.9\% (+0.3\% over BERT, +0.8\% over HateBERT), \texttt{OwsDeu} a German average of 75.0\% (+3.1\% over BERT, +5.7\% over HateBERT), and \texttt{OwsSpa} a Spanish average of 75.5\% (+0.5\% over BERT, +3.0\% over HateBERT).
HateBERT's English Reddit pre-training offers no advantage on German or Spanish, explaining the large OWS margins in those languages.

\paragraph{Cross-language effects and overall average.}
OWS pre-training also produces cross-lingual gains: \texttt{OwsDeu}, trained on only German, boosts English-test performance by +11.2\% over BERT and Vietnamese by +6.4\%, likely because the German OWS crawl contains multilingual web content.
On the 16-set overall average, \texttt{OwsDeu} (66.9\%) and \texttt{OwsEng} (72.3\%) both surpass BERT and HateBERT.
The sole exception is \texttt{OwsSpa} (43.6\%), which leads on its three Spanish benchmarks but collapses on the thirteen non-Spanish test sets because neither its pre-training data nor its fine-tuning set contains other languages.

\paragraph{Where HateBERT outperforms OWS models.}
HateBERT's advantage is limited to one scenario: when fine-tuned on non-English data only, it retains strong English performance from its Reddit pre-training (e.g., English average of 55.4\% under the Spanish config versus 35.5\% for \texttt{OwsSpa}).
In all other configurations the OWS models match or exceed HateBERT.

\begin{table*}[t]
    \begin{subtable}[t]{\linewidth}
    \centering
    \caption*{BERT Models}
    \resizebox{\linewidth}{!}{
    \begin{tabular}{|l|ccc|ccc|ccc|ccc|ccc|ccc|} 
    \toprule
    \textbf{Train Set} &
    \multicolumn{3}{c|}{\textbf{Spanish (Spa)}} &
    \multicolumn{3}{c|}{\textbf{German (Deu)}} &
    \multicolumn{3}{c|}{\textbf{English (Eng)}} &
    \multicolumn{3}{c|}{\textbf{7-Set Mix}} &
    \multicolumn{3}{c|}{\textbf{7-Set Mix + Synth.}} &
    \multicolumn{3}{c}{\textbf{16-Set Mix}} \\
    \cmidrule(lr){2-4} \cmidrule(lr){5-7} \cmidrule(lr){8-10} \cmidrule(lr){11-13} \cmidrule(lr){14-16}\cmidrule(lr){17-19}
    \textbf{Models} &
    \textcolor{NavyBlue}{\textbf{BERT}} &
    \textcolor{DarkAmber}{\textbf{HateBERT}} &
    \textcolor{ModernPurple}{\textbf{OwsSpa}} &
    \textcolor{NavyBlue}{\textbf{BERT}} &
    \textcolor{DarkAmber}{\textbf{HateBERT}} &
    \textcolor{ModernPurple}{\textbf{OwsDeu}} &
    \textcolor{NavyBlue}{\textbf{BERT}} &
    \textcolor{DarkAmber}{\textbf{HateBERT}} &
    \textcolor{ModernPurple}{\textbf{OwsEng}} &
    \textcolor{NavyBlue}{\textbf{BERT}} &
    \textcolor{DarkAmber}{\textbf{HateBERT}} &
    \textcolor{ModernPurple}{\textbf{Ows4L}} &
    \textcolor{NavyBlue}{\textbf{BERT}} &
    \textcolor{DarkAmber}{\textbf{HateBERT}} &
    \textcolor{ModernPurple}{\textbf{Ows4L}} &
    \textcolor{NavyBlue}{\textbf{BERT}} &
    \textcolor{DarkAmber}{\textbf{HateBERT}} &
    \textcolor{ModernPurple}{\textbf{Ows4L}} \\
    \midrule
    \rowcolor{RoyalBlue!8}
    \textcolor{RoyalBlue!80!black}{HateXplain} & 30.3 & 54.7 & 28.9 & 52.6 & 55.5 & 59.7 & \cellcolor{RoyalBlue!28}77.7 & \cellcolor{RoyalBlue!28}77.5 & \cellcolor{RoyalBlue!28}77.9 & 77.7 & 77.6 & 77.6 & \textbf{\underline{78.9}} & 77.8 & 78.4 & 78.0 & 77.1 & 77.5 \\
    \rowcolor{RoyalBlue!8}
    \textcolor{RoyalBlue!80!black}{Sexism} & 47.7 & 49.9 & 46.4 & 51.8 & 56.6 & 56.5 & \cellcolor{RoyalBlue!28}85.1 & \cellcolor{RoyalBlue!28}84.9 & \cellcolor{RoyalBlue!28}\textbf{\underline{86.7}} & 85.9 & 85.8 & 85.4 & 85.2 & 84.9 & 85.1 & 84.5 & 85.3 & 86.4 \\
    \rowcolor{RoyalBlue!8}
    \textcolor{RoyalBlue!80!black}{Covid} & 44.6 & 48.2 & 44.6 & 54.6 & 52.9 & 63.1 & \cellcolor{RoyalBlue!28}77.5 & \cellcolor{RoyalBlue!28}75.2 & \cellcolor{RoyalBlue!28}79.5 & \textbf{\underline{81.1}} & 79.6 & 81.0 & 77.4 & 76.8 & 76.9 & 78.1 & 76.3 & 78.8 \\
    \rowcolor{RoyalBlue!8}
    \textcolor{RoyalBlue!80!black}{US\_election} & 46.6 & 47.9 & 47.4 & 46.6 & 51.1 & 46.6 & \cellcolor{RoyalBlue!28}66.5 & \cellcolor{RoyalBlue!28}67.7 & \cellcolor{RoyalBlue!28}\textbf{\underline{70.4}} & 64.2 & 63.2 & 68.8 & 64.4 & 62.4 & 64.7 & 67.5 & 68.5 & 68.7 \\
    \rowcolor{RoyalBlue!8}
    \textcolor{RoyalBlue!80!black}{HateEval-eng} & 37.4 & 54.0 & 36.4 & 47.2 & 56.4 & 58.8 & \cellcolor{RoyalBlue!28}\textbf{\underline{77.3}} & \cellcolor{RoyalBlue!28}75.6 & \cellcolor{RoyalBlue!28}75.9 & 58.7 & 60.5 & 60.0 & 60.2 & 62.2 & 61.4 & 77.0 & 76.3 & 76.2 \\
    \rowcolor{RoyalBlue!8}
    \textcolor{RoyalBlue!80!black}{AbusEval} & 44.2 & 53.5 & 44.2 & 48.6 & 58.6 & 51.4 & \cellcolor{RoyalBlue!28}67.0 & \cellcolor{RoyalBlue!28}67.1 & \cellcolor{RoyalBlue!28}70.7 & 52.1 & 55.1 & 52.5 & 51.7 & 56.7 & 55.9 & 68.6 & 67.1 & \textbf{\underline{72.1}} \\
    \rowcolor{RoyalBlue!8}
    \textcolor{RoyalBlue!80!black}{AHSD} & 16.1 & 39.5 & 14.4 & 28.8 & 48.5 & 45.3 & \cellcolor{RoyalBlue!28}91.7 & \cellcolor{RoyalBlue!28}90.6 & \cellcolor{RoyalBlue!28}91.7 & 49.4 & 50.6 & 54.0 & 78.4 & 79.1 & 79.4 & 91.4 & 91.0 & \textbf{\underline{91.9}} \\
    \rowcolor{RoyalBlue!28}
    \textit{Eng Avg} & 36.8 & 55.4 & 35.5\,{\scriptsize\textcolor{BrickRed}{-1.3}} & 52.7 & 61.4 & 63.9\,{\scriptsize\textcolor{ForestGreen}{+11.2}} & 86.6 & 86.1 & 86.9\,{\scriptsize\textcolor{ForestGreen}{+0.3}} & 74.6 & 75.0 & 76.2\,{\scriptsize\textcolor{ForestGreen}{+1.6}} & 83.1 & 83.0 & 83.2\,{\scriptsize\textcolor{ForestGreen}{+0.1}} & 86.6 & 86.2 & 86.8\,{\scriptsize\textcolor{ForestGreen}{+0.2}} \\
    \midrule
    \rowcolor{SeaGreen!8}
    \textcolor{SeaGreen!80!black}{GermEval21} & 39.4 & 38.9 & 38.8 & \cellcolor{SeaGreen!28}59.9 & \cellcolor{SeaGreen!28}58.4 & \cellcolor{SeaGreen!28}\textbf{\underline{60.3}} & 39.3 & 39.9 & 39.2 & 56.4 & 57.5 & 58.1 & 51.5 & 54.7 & 54.4 & 59.8 & 59.4 & 60.0 \\
    \rowcolor{SeaGreen!8}
    \textcolor{SeaGreen!80!black}{GermEval19} & 40.0 & 39.9 & 39.9 & \cellcolor{SeaGreen!28}74.8 & \cellcolor{SeaGreen!28}70.8 & \cellcolor{SeaGreen!28}78.8 & 41.7 & 41.7 & 41.3 & 73.4 & 69.8 & 74.6 & 72.0 & 71.0 & 73.5 & 77.5 & 73.4 & \textbf{\underline{79.1}} \\
    \rowcolor{SeaGreen!8}
    \textcolor{SeaGreen!80!black}{GermEval18} & 40.1 & 39.8 & 39.7 & \cellcolor{SeaGreen!28}76.4 & \cellcolor{SeaGreen!28}72.8 & \cellcolor{SeaGreen!28}81.4 & 42.0 & 42.4 & 41.0 & 80.5 & 75.5 & \textbf{\underline{83.1}} & 76.3 & 74.6 & 78.6 & 80.5 & 75.7 & 82.6 \\
    \rowcolor{SeaGreen!8}
    \textcolor{SeaGreen!80!black}{HASOC} & 42.6 & 42.7 & 42.7 & \cellcolor{SeaGreen!28}73.5 & \cellcolor{SeaGreen!28}71.3 & \cellcolor{SeaGreen!28}72.5 & 47.2 & 48.3 & 46.3 & 68.0 & 68.5 & 69.3 & 69.1 & 67.7 & 71.9 & \textbf{\underline{75.4}} & 73.7 & 74.8 \\
    \rowcolor{SeaGreen!8}
    \textcolor{SeaGreen!80!black}{Gahd} & 37.7 & 36.5 & 36.9 & \cellcolor{SeaGreen!28}71.0 & \cellcolor{SeaGreen!28}70.0 & \cellcolor{SeaGreen!28}\textbf{\underline{73.2}} & 39.2 & 40.8 & 38.9 & 53.2 & 53.9 & 55.5 & 45.7 & 46.5 & 45.9 & 72.9 & 71.2 & \textbf{\underline{73.2}} \\
    \rowcolor{SeaGreen!28}
    \textit{Deu Avg} & 39.7 & 39.2 & 39.2\,{\scriptsize\textcolor{BrickRed}{-0.5}} & 71.9 & 69.3 & 75.0\,{\scriptsize\textcolor{ForestGreen}{+3.1}} & 41.1 & 41.8 & 40.7\,{\scriptsize\textcolor{BrickRed}{-0.4}} & 68.3 & 66.0 & 70.2\,{\scriptsize\textcolor{ForestGreen}{+1.9}} & 64.8 & 64.6 & 66.6\,{\scriptsize\textcolor{ForestGreen}{+1.8}} & 74.4 & 71.3 & 75.5\,{\scriptsize\textcolor{ForestGreen}{+1.1}} \\
    \midrule
    \rowcolor{BurntOrange!8}
    \textcolor{BurntOrange!90!black}{ViHSD} & 45.0 & 45.7 & 45.0 & 49.1 & 50.9 & 55.5 & 45.2 & 45.6 & 45.0 & 72.2 & 67.4 & \textbf{\underline{74.0}} & 71.0 & 64.7 & 72.8 & 72.3 & 67.9 & 73.8 \\
    \midrule
    \rowcolor{Orchid!8}
    \textcolor{Orchid!80!black}{Haternet} & \cellcolor{Orchid!28}67.3 & \cellcolor{Orchid!28}64.9 & \cellcolor{Orchid!28}68.1 & 45.7 & 48.7 & 56.8 & 43.0 & 45.1 & 44.4 & 45.2 & 46.7 & 48.0 & 47.2 & 45.2 & 49.1 & 69.3 & 67.7 & \textbf{\underline{71.9}} \\
    \rowcolor{Orchid!8}
    \textcolor{Orchid!80!black}{HateEval-spa} & \cellcolor{Orchid!28}75.9 & \cellcolor{Orchid!28}73.5 & \cellcolor{Orchid!28}76.6 & 38.3 & 38.4 & 46.9 & 36.2 & 36.7 & 36.3 & 36.9 & 36.7 & 38.6 & 38.5 & 36.5 & 38.2 & 78.6 & 76.7 & \textbf{\underline{79.0}} \\
    \rowcolor{Orchid!8}
    \textcolor{Orchid!80!black}{Chileno} & \cellcolor{Orchid!28}55.4 & \cellcolor{Orchid!28}49.4 & \cellcolor{Orchid!28}56.8 & 50.1 & 49.7 & 48.8 & 48.5 & 48.5 & 48.5 & 50.2 & 49.0 & 50.2 & 49.8 & 49.1 & 49.6 & \textbf{\underline{57.1}} & 54.0 & 56.0 \\
    \rowcolor{Orchid!28}
    \textit{Spa Avg} & 75.0 & 72.5 & 75.5\,{\scriptsize\textcolor{ForestGreen}{+0.5}} & 46.0 & 46.8 & 52.9\,{\scriptsize\textcolor{ForestGreen}{+6.9}} & 44.0 & 44.9 & 44.4\,{\scriptsize\textcolor{ForestGreen}{+0.4}} & 45.2 & 45.5 & 47.1\,{\scriptsize\textcolor{ForestGreen}{+1.9}} & 46.7 & 44.9 & 47.3\,{\scriptsize\textcolor{ForestGreen}{+0.6}} & 76.9 & 75.1 & 77.4\,{\scriptsize\textcolor{ForestGreen}{+0.5}} \\
    \midrule
    \textbf{Avg} & 44.1 & 57.2 & 43.6\,{\scriptsize\textcolor{BrickRed}{-0.5}} & 61.3 & 65.0 & 66.9\,{\scriptsize\textcolor{ForestGreen}{+5.6}} & 72.1 & 71.9 & 72.3\,{\scriptsize\textcolor{ForestGreen}{+0.2}} & 71.6 & 70.7 & 73.1\,{\scriptsize\textcolor{ForestGreen}{+1.5}} & 75.2 & 74.7 & 75.7\,{\scriptsize\textcolor{ForestGreen}{+0.5}} & 81.5 & 79.9 & \textbf{\underline{82.1}}\,{\scriptsize\textcolor{ForestGreen}{+0.6}} \\
    \bottomrule
    \end{tabular}
    }
    \end{subtable}
    \caption{Macro-F1 (\%) for \textcolor{NavyBlue}{\textbf{BERT}}, \textcolor{DarkAmber}{\textbf{HateBERT}}, and four \textcolor{ModernPurple}{\textbf{OWS-BERT}} variants across six training configurations.
    Bold-underlined = best per row.
    Darker-shaded cells = \emph{in-domain} results (training language matches test language).
    Per-language averages show the OWS\,--\,BERT delta ($\Delta$) on the OWS column.}
    \label{tab:BertCompare}
\end{table*}

\subsubsection{Multilingual Continued Pre-Training}

\texttt{Ows4L} outperforms both BERT and HateBERT on all four per-language averages and on the overall average under every multilingual training configuration.
Under \emph{7-Set Mix}, the gain over BERT is +1.6\% (Eng), +1.9\% (Deu), +1.8\% (ViHSD), and +1.9\% (Spa), with an overall average of 73.1\% versus 71.6\% for BERT and 70.7\% for HateBERT.
The benefit is most pronounced in low-data and low-resource settings: ViHSD improves by +6.6\% over HateBERT, and AHSD by +3.4\%.
Under \emph{16-Mix}, the gap narrows but \texttt{Ows4L} still leads (82.1\% versus 81.5\% BERT and 79.9\% HateBERT), achieving the best score on 6 of 16 test sets.
These results indicate that OWS continued pre-training is \emph{most valuable when supervised data is limited}.

\paragraph{Benefit of adding synthetic data.}
Comparing \emph{7-Set + Synth.} against \emph{7-Set Mix} isolates the effect of the ${\approx}$46K \texttt{Qwen2.5-14B}-annotated OWS texts.
All three models improve on the overall average: BERT rises from 71.6\% to 75.2\% (+3.6\%), HateBERT from 70.7\% to 74.7\% (+4.0\%), and \texttt{Ows4L} from 73.1\% to 75.7\% (+2.6\%).
The gains concentrate on English (Eng Avg: +7.0\% to +8.5\% across models), whereas German regresses ($-1.4$\% to $-3.6$\%) and Vietnamese drops slightly ($-1.2$\% to $-2.7$\%).
Notably, $\texttt{Ows4L}$ still leads under this setting (75.7\% versus 75.2\% BERT and 74.7\% HateBERT), but the gap narrows compared to 7-Set Mix, suggesting that synthetic data partially compensates for the advantage that OWS continued pre-training provides.

\subsection{RQ2: LLM Ensemble Annotations}
\label{sec:rq2}

\subsubsection{Small-Scale Models: Llama3.2-1B and BERT}

Table~\ref{tab:1B} shows that synthetic data substantially benefits smaller models.
\texttt{Llama3.2-1B} gains 2--11\% pooled F1 over the human baseline depending on the ensemble strategy.
The strongest configuration is \texttt{H+LGB} (human + LGB-labelled data), reaching 65.4\% overall F1 (+10.6\%), with per-dataset gains up to +25\% on AHSD and +19\% on Gahd (Table~\ref{tab:1B-full}).

\paragraph{English and German.}
For the three synthetic-only strategies, English and German pooled F1 consistently exceeds the human baseline: up to +8\% for English and +22\% for German, with \texttt{LGB} yielding the largest margins in both languages.

\paragraph{Vietnamese.}
All three synthetic-only strategies slightly trail the human baseline at the pooled level (F1 58--59 vs.\ 59.3), though \texttt{H+LGB} recovers strongly (+6.8\%).
Vietnamese comprises only ${\approx}$2.8\% of the synthetic corpus (${\approx}$6,600 of 240K texts); the limited volume constrains standalone LLM annotation quality.

\paragraph{Spanish (zero-shot).}
No training configuration includes Spanish data.
At the pooled level, \texttt{Mean} and \texttt{Vote} underperform the human baseline by 10--11\% F1, because the synthetic corpus contains no Spanish texts and equal-weight aggregation cannot compensate.
By contrast, \texttt{LGB} is the only synthetic-only strategy to exceed the human baseline (+1.6\% pooled F1), confirming that the meta-learner acquires a degree of language-agnostic calibration.

\paragraph{Why LightGBM outperforms Vote and Mean.}
\texttt{Vote} and \texttt{Mean} treat all four annotators equally, amplifying any systematic biases shared across them.
\texttt{LGB} learns \emph{which} LLMs are reliable on \emph{which} patterns and discovers decision boundaries that align with human annotations.
The result is fewer false negatives in low-resource languages and better-calibrated probabilities overall.

\begin{table}[t]
\centering
\footnotesize
\setlength{\tabcolsep}{3pt}
\begin{tabular}{@{}l c rrrrrr@{}}
\toprule
 & \textbf{\#} & \textbf{Base} & \textbf{Hum.} & \textbf{Mean} & \textbf{Vote} & \textbf{LGB} & \textbf{H+LGB} \\
\midrule
\multicolumn{8}{@{}l}{\textit{Llama3.2-1B}} \\
\rowcolor{RoyalBlue!8} EN  & 7 & 51.7 & 59.7 & 61.5 & 61.9 & 67.3 & \textbf{75.4} \\
\rowcolor{SeaGreen!8}  DE  & 5 & 44.6 & 42.7 & 55.6 & 54.3 & \textbf{64.4} & 57.8 \\
\rowcolor{BurntOrange!8} VI & 1 & 47.1 & 59.3 & 58.5 & 57.6 & 58.3 & \textbf{66.1} \\
\rowcolor{Orchid!8}     ES & 3 & 33.0 & 46.5 & 36.7 & 35.2 & 48.1 & \textbf{47.4} \\
\cmidrule(lr){1-8}
\rowcolor{NavyBlue!6}  7-Set & 7 & 50.2 & 60.2 & 62.6 & 62.2 & 66.9 & \textbf{69.7} \\
\rowcolor{NavyBlue!6}  Rest  & 9 & 43.0 & 48.2 & 49.7 & 49.2 & 59.1 & \textbf{60.3} \\
\rowcolor{gray!15}     \textbf{All} & 16 & 47.4 & 54.8 & 56.9 & 56.5 & 63.1 & \textbf{65.4} \\
\midrule
\multicolumn{8}{@{}l}{\textit{Qwen2.5-14B}} \\
\rowcolor{RoyalBlue!8} EN  & 7 & 69.4 & \textbf{78.6} & 78.5 & 74.2 & 76.5 & \textbf{79.4} \\
\rowcolor{SeaGreen!8}  DE  & 5 & 71.7 & 77.0 & 75.0 & 69.9 & 74.9 & \textbf{76.1} \\
\rowcolor{BurntOrange!8} VI & 1 & 71.9 & 76.0 & 76.0 & 70.9 & 75.9 & \textbf{77.5} \\
\rowcolor{Orchid!8}     ES & 3 & 43.7 & 61.3 & 62.7 & \textbf{67.9} & 64.3 & 63.9 \\
\cmidrule(lr){1-8}
\rowcolor{NavyBlue!6}  7-Set & 7 & 67.4 & 76.6 & 76.0 & 73.8 & 76.0 & \textbf{77.2} \\
\rowcolor{NavyBlue!6}  Rest  & 9 & 67.6 & 75.3 & 75.3 & 71.5 & 73.8 & \textbf{75.7} \\
\rowcolor{gray!15}     \textbf{All} & 16 & 67.8 & 76.1 & 75.9 & 72.8 & 75.1 & \textbf{76.7} \\
\bottomrule
\end{tabular}
\caption{Pooled macro-F1~(\%) by language and training split for LoRA-tuned LLMs.
\textbf{\#}~= number of test sets; \textbf{7-Set}/\textbf{Rest}~= training vs.\ held-out datasets.
Scores are computed on the pooled predictions within each group, not averaged across datasets.
The classification threshold is the mean predicted probability of the \textit{Hate} class, not the default~0.5.
Full per-dataset breakdown in Table~\ref{tab:1B-full} (Appendix).}
\label{tab:1B}
\end{table}

\subsubsection{Large-Scale Model: Qwen2.5-14B}

\texttt{Qwen2.5-14B} fine-tuned on the 7-Set already approaches the best BERT models trained on all sixteen sets.
Adding synthetic data yields a modest +0.6\% pooled F1 with \texttt{H+LGB} (76.7 vs.\ 76.1).
Language-level trends diverge:
\begin{itemize}[leftmargin=*,nosep]
    \item \textbf{English (+0.8\%):} Small but consistent gains; per-dataset highlights include Covid (+5\% F1) and HateEval-eng (+2.5\%) (Table~\ref{tab:1B-full}).
    \item \textbf{Spanish (+2.6\%):} The clearest cross-lingual benefit---despite no Spanish training data, all ensemble strategies improve pooled F1, with \texttt{Vote} reaching +6.6\%.
    \item \textbf{German ($-$0.9\%):} Pooled F1 dips slightly; under \texttt{Vote} the drop reaches $-$7.1\%.
    The model already achieves high German F1 on human data (77.0); the imbalanced synthetic set (${<}$3\% hate) shifts its decision boundary towards \textit{Neutral}, and \texttt{Vote} amplifies correlated LLM errors on German text.
\end{itemize}

\section{Discussion}

\subsection{Answers to Research Questions}

\paragraph{RQ1: Value of unlabelled web data.}
OWS continued pre-training reliably improves BERT-family models, especially in multilingual low-data settings.
The effect is largest for \texttt{Ows4L} when training data is scarce (+3\% average F1 under 7-Set) and diminishes as supervised data grows (+1\% under 16-Mix).
Neither monolingual OWS models nor HateBERT provide broad cross-lingual generalisation; only \texttt{Ows4L} achieves consistent gains across all four language groups.

\paragraph{RQ2: LLM ensemble annotations.}
The value of synthetic annotations depends strongly on (a)~downstream model capacity and (b)~language coverage of the synthetic set.
For \texttt{Llama3.2-1B}, \texttt{H+LGB} labels yield +10.6\% pooled F1 over the human baseline, effectively distilling knowledge from larger annotation LLMs into a small downstream model.
For \texttt{Qwen2.5-14B}, gains are modest (+0.6\% pooled F1); its strong priors quickly saturate on the imbalanced synthetic data.
The LightGBM meta-learner is the most effective strategy because it learns model-reliability weights from human labels, making it robust to the shared biases of individual LLMs.

\subsection{Limitations}
\label{sec:limitations}

\paragraph{Corpus quality.}
The URL/schema-based OWS filter is efficient but cannot guarantee conversational content quality.
Manual inspection of a random sample and more precise classifier-based filtering would yield a cleaner pre-training corpus.

\paragraph{Label space collapse.}
Mapping hate, offensive, and abusive labels to a single positive class may conflate qualitatively different phenomena and reduce generalisation to datasets that maintain fine-grained distinctions.

\paragraph{LLM annotation bias.}
Synthetic labels inherit biases from the four annotating LLMs, all predominantly English-centric.
The under-performance of \texttt{Mean} and \texttt{Vote} on Vietnamese and Spanish reflects this.

\paragraph{Ethical considerations.}
During data collection, no personally identifiable information (usernames, author profiles, or metadata) was gathered; we retained only raw texts, comments, and posts from publicly indexed web pages.

\subsection{Future Work}

We plan to
(i)~scale the OWS corpus substantially, increasing both volume and language diversity, to strengthen continued pre-training;
(ii)~adopt more capable recent SLMs and LLMs to assess whether newer architectures widen the gap between small and large models;
(iii)~extend coverage to additional low-resource languages;
(iv)~explore improved prompting strategies (chain-of-thought, calibration prompts) to reduce annotation bias and label imbalance;
and (v)~investigate hate-focused corpus filtering to increase the proportion of relevant content in OWS samples.

\section{Conclusion}

We presented a large-scale benchmark study examining how unlabelled web data and ensemble LLM annotations can improve multilingual hate-speech detection across sixteen benchmarks in four languages.
Our key findings are:
\begin{enumerate}[leftmargin=*,nosep]
    \item Domain-adaptive continued pre-training on OWS data provides consistent F1 gains, with the multilingual \texttt{Ows4L} achieving the best average macro-F1 (77.0\%) across all configurations.
    \item The LightGBM ensemble outperforms mean averaging and majority voting and is the only synthetic strategy that reliably avoids regressions on unseen languages.
    \item The benefit of synthetic data scales inversely with model capacity: +10.6\% pooled F1 for \texttt{Llama3.2-1B}, but only +0.6\% for \texttt{Qwen2.5-14B}.
    \item Severe class imbalance in OWS-derived synthetic sets remains a critical bottleneck, especially for low-resource languages.
\end{enumerate}

\noindent
These findings motivate future work on scaling OWS data, adopting updated models, and developing richer annotation strategies.

\section*{Acknowledgements}
This work has received funding from the Bavarian State Ministry of Economic Affairs, Regional Development, and Energy (StMWi).

\vspace{4pt}
\noindent\includegraphics[height=1.6cm]{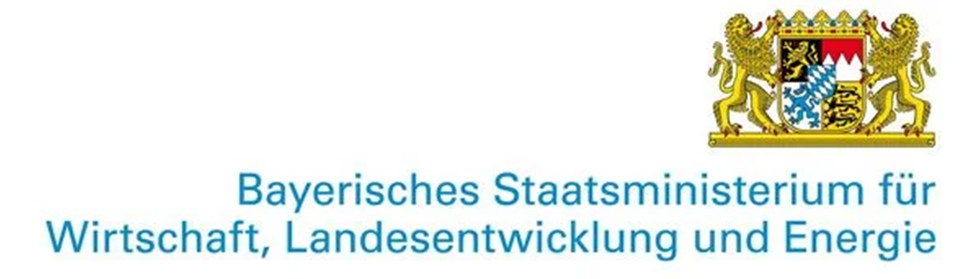}

\section{References}\label{sec:reference}

\bibliographystyle{lrec2026-natbib}
\bibliography{cameraReady}

\bibliographystylelanguageresource{lrec2026-natbib}
\bibliographylanguageresource{languageresource}

\appendix
\section{LLM Annotation Details}\label{app:annotation-details}

\subsection{Annotation Prompt}\label{app:prompt}

Each OWS text was classified by the four instruction-tuned LLMs using the following zero-shot prompt:

\begin{quote}
\small\ttfamily\raggedright
You are tasked with annotating speech.
Your response must be a single valid number:\par
1 for Hate/Offensive/Sexism/Toxic/Political/\linebreak
COVID-related Hate Speech,\par
2 for Neutral Speech.\par\medskip
Provide only the number corresponding to the
category. Do not include any explanation or
additional text.\par
Do you think the following comment is
Hate/Offensive/Sexism/Toxic/Political/\linebreak
COVID-related Hate Speech, speech or Neutral
speech?\par\medskip
"\{comment\}"\par\medskip
Your Answer:
\end{quote}

Token-level probabilities for the target labels were extracted from the output distribution of each model.
For the 240K\,ensemble annotation pool (Section~\ref{sec:llm-annotation}), all four models were queried; for the 46K\,BERT augmentation subset (7-Set\,+\,Synth.), only \texttt{Qwen2.5-14B} was used.

\subsection{LightGBM Meta-Learner Configuration}\label{app:lgb-config}

The LightGBM meta-learner (\texttt{LGB}) was trained with the following hyperparameters:

\begin{table}[h]
\centering
\small
\caption{LightGBM training configuration for the ensemble meta-learner.}
\label{tab:lgb-config}
\begin{tabular}{ll}
\toprule
\textbf{Parameter} & \textbf{Value} \\
\midrule
objective        & binary \\
metric           & binary\_logloss \\
boosting\_type   & gbdt \\
num\_leaves      & 34 \\
learning\_rate   & 0.05 \\
feature\_fraction & 0.9 \\
bagging\_fraction & 0.8 \\
bagging\_freq    & 5 \\
\bottomrule
\end{tabular}
\end{table}

Two separate binary classifiers were trained---one for the Hate class and one for the Neutral class---using the eight-dimensional probability vectors (two classes $\times$ four models) from the human-labelled training sets as features.

\subsection{240K Ensemble Annotation Statistics}\label{app:240k-stats}

Table~\ref{tab:240k-stats} summarises the 240,647-text annotation pool used for LLM ensemble labelling.

\begin{table}[h]
\centering
\small
\caption{Per-model and ensemble statistics of the 240K annotation pool, broken down by language.
  ``\%\,Hate'' = fraction of texts with $P(\mathrm{Hate})>0.5$ (per-model) or positive label (ensemble).}
\label{tab:240k-stats}
\resizebox{\linewidth}{!}{
\begin{tabular}{lrrr|r}
\toprule
 & \textbf{deu} & \textbf{eng} & \textbf{vie} & \textbf{All} \\
 & 125\,617 & 108\,375 & 6\,655 & 240\,647 \\
\midrule
\multicolumn{5}{l}{\textit{Per-model mean $P(\mathrm{Hate})$}} \\
\quad Qwen2.5-14B   & 0.027 & 0.021 & 0.031 & 0.024 \\
\quad Gemma2-9B      & 0.107 & 0.039 & 0.113 & 0.080 \\
\quad Llama3.1-8B    & 0.002 & 0.001 & 0.001 & 0.011 \\
\quad Mistral-7B     & 0.026 & 0.015 & 0.023 & 0.021 \\
\midrule
\multicolumn{5}{l}{\textit{Per-model \% classified Hate ($P>0.5$)}} \\
\quad Qwen2.5-14B   & 2.69 & 1.96 & 3.56 & 2.38 \\
\quad Gemma2-9B      & 8.41 & 3.25 & 6.40 & 6.03 \\
\quad Llama3.1-8B    & 0.21 & 0.19 & 0.03 & 0.19 \\
\quad Mistral-7B     & 2.43 & 1.60 & 2.48 & 2.05 \\
\midrule
\multicolumn{5}{l}{\textit{Ensemble \% Hate}} \\
\quad Mean            & 1.98 & 1.27 & 1.89 & 1.66 \\
\quad LGB             & 1.57 & 0.94 & 2.01 & 1.30 \\
\quad Vote            & 2.37 & 1.45 & 2.42 & 1.96 \\
\bottomrule
\end{tabular}
}
\end{table}

\subsection{46K BERT Augmentation Statistics}\label{app:46k-stats}

Table~\ref{tab:46k-stats} summarises the 45,940-text subset annotated solely by \texttt{Qwen2.5-14B} and used as synthetic training data for the BERT 7-Set\,+\,Synth.\ configuration.

\begin{table}[h]
\centering
\small
\caption{Label distribution of the 46K BERT augmentation set, annotated by \texttt{Qwen2.5-14B}, broken down by language.}
\label{tab:46k-stats}
\resizebox{\linewidth}{!}{
\begin{tabular}{lrrrr|r}
\toprule
 & \textbf{deu} & \textbf{eng} & \textbf{vie} & \textbf{spa} & \textbf{All} \\
 & 26\,084 & 11\,635 & 6\,871 & 1\,350 & 45\,940 \\
\midrule
\multicolumn{6}{l}{\textit{Qwen2.5-14B label counts}} \\
\quad Neutral    & 22\,702 & 4\,043 & 6\,341 & 1\,143 & 34\,229 \\
\quad Hate       & 3\,284  & 3\,943 & 474    & 149    & 7\,850 \\
\quad Offensive  & 98      & 3\,649 & 56     & 58     & 3\,861 \\
\midrule
\multicolumn{6}{l}{\textit{Qwen2.5-14B label \%}} \\
\quad Neutral    & 87.03 & 34.75 & 92.29 & 84.67 & 74.51 \\
\quad Hate       & 12.59 & 33.89 & 6.90  & 11.04 & 17.09 \\
\quad Offensive  & 0.38  & 31.36 & 0.82  & 4.30  & 8.40 \\
\midrule
\multicolumn{6}{l}{\textit{Mean $P(\mathrm{Hate})$}} \\
\quad Qwen2.5-14B & \multicolumn{4}{c}{---} & 0.451 \\
\quad Gemma2-9B   & \multicolumn{4}{c}{---} & 0.527 \\
\bottomrule
\end{tabular}
}
\end{table}

\subsection{OWS Schema Types}\label{app:ows-schema}

The following ten \texttt{schema.org} types (matched under both \texttt{http} and \texttt{https} schemes) were used to filter the OpenWebIndex for conversational and user-generated content:

\begin{enumerate}\itemsep0pt\parsep0pt
  \item \texttt{DiscussionForumPosting}
  \item \texttt{SocialMediaPosting}
  \item \texttt{BlogPosting}
  \item \texttt{Article}
  \item \texttt{Comment}
  \item \texttt{UserComments}
  \item \texttt{QAPage}
  \item \texttt{Question}
  \item \texttt{Review}
  \item \texttt{Blog}
\end{enumerate}

\section{Full Per-Dataset LLM Results}\label{app:llm-full}

\begin{table*}[ht]
    \begin{subtable}[t]{\linewidth}
    \centering
    \caption*{\texttt{Llama3.2-1B}}
    \resizebox{\linewidth}{!}{
    \begin{tabular}{|lcc|cc|cc|cc|cc|cc|} 
        \toprule
        \textbf{Dataset} & \multicolumn{2}{c|}{\textbf{Base}} &
        \multicolumn{2}{c|}{\textbf{Human}} &
        \multicolumn{2}{c|}{\textbf{Mean}} &
        \multicolumn{2}{c|}{\textbf{Voting}} &
        \multicolumn{2}{c|}{\textbf{LGB}} &
        \multicolumn{2}{c|}{\textbf{Human+LGB}} \\
        \cmidrule(lr){2-3} \cmidrule(lr){4-5} \cmidrule(lr){6-7} \cmidrule(lr){8-9} \cmidrule(lr){10-11} \cmidrule(lr){12-13}
        &  Acc & F1 &   Acc & F1 &  Acc & F1 &  Acc & F1  & Acc & F1  & Acc & F1 \\
        \midrule
   \textcolor{RoyalBlue!80!black}{\textbf{HateXplain}} & 53.6 & 53.3 & 58.4 & 58.4 & 60.8 {\scriptsize \textcolor{ForestGreen}{+02.4}} & 60.7 {\scriptsize \textcolor{ForestGreen}{+02.3}} & 61.3 {\scriptsize \textcolor{ForestGreen}{+02.9}} & 61.3 {\scriptsize \textcolor{ForestGreen}{+02.9}} & 67.7 {\scriptsize \textcolor{ForestGreen}{+09.3}} & 67.0 {\scriptsize \textcolor{ForestGreen}{+08.6}} & 70.4 {\scriptsize \textcolor{ForestGreen}{+12.0}} & \textbf{67.3} {\scriptsize \textcolor{ForestGreen}{+08.9}} \\
  \textcolor{RoyalBlue!80!black}{\textbf{Sexism}} & 74.2 & 47.5 & 78.5 & 59.8 & 82.3 {\scriptsize \textcolor{ForestGreen}{+03.8}} & 51.7 {\scriptsize \textcolor{BrickRed}{-08.1}} & 81.1 {\scriptsize \textcolor{ForestGreen}{+02.6}} & 50.9 {\scriptsize \textcolor{BrickRed}{-08.9}} & 82.6 {\scriptsize \textcolor{ForestGreen}{+04.1}} & 56.3 {\scriptsize \textcolor{BrickRed}{-03.5}} & 83.5 {\scriptsize \textcolor{ForestGreen}{+05.0}} & \textbf{68.5} {\scriptsize \textcolor{ForestGreen}{+08.7}} \\
  \textcolor{RoyalBlue!80!black}{\textbf{Covid}} & 38.2 & 37.7 & 75.0 & 63.0 & 77.3 {\scriptsize \textcolor{ForestGreen}{+02.3}} & 65.1 {\scriptsize \textcolor{ForestGreen}{+02.1}} & 76.1 {\scriptsize \textcolor{ForestGreen}{+01.1}} & 64.6 {\scriptsize \textcolor{ForestGreen}{+01.6}} & 69.6 {\scriptsize \textcolor{BrickRed}{-05.4}} & 62.7 {\scriptsize \textcolor{BrickRed}{-00.3}} & 76.3 {\scriptsize \textcolor{ForestGreen}{+01.3}} & \textbf{69.6} {\scriptsize \textcolor{ForestGreen}{+06.6}} \\
  \textcolor{RoyalBlue!80!black}{\textbf{US\_election}} & 30.3 & 29.9 & 83.4 & 51.6 & 82.3 {\scriptsize \textcolor{BrickRed}{-01.1}} & 59.8 {\scriptsize \textcolor{ForestGreen}{+08.2}} & 81.7 {\scriptsize \textcolor{BrickRed}{-01.7}} & 60.0 {\scriptsize \textcolor{ForestGreen}{+08.4}} & 79.9 {\scriptsize \textcolor{BrickRed}{-03.5}} & \textbf{62.3} {\scriptsize \textcolor{ForestGreen}{+10.7}} & 85.2 {\scriptsize \textcolor{ForestGreen}{+01.8}} & 56.8 {\scriptsize \textcolor{ForestGreen}{+05.2}} \\
  \textcolor{RoyalBlue!80!black}{HateEval-eng} & 49.5 & 49.5 & 58.2 & 51.5 & 61.9 {\scriptsize \textcolor{ForestGreen}{+03.7}} & 56.8 {\scriptsize \textcolor{ForestGreen}{+05.3}} & 62.1 {\scriptsize \textcolor{ForestGreen}{+03.9}} & 57.6 {\scriptsize \textcolor{ForestGreen}{+06.1}} & 64.0 {\scriptsize \textcolor{ForestGreen}{+05.8}} & 62.1 {\scriptsize \textcolor{ForestGreen}{+10.6}} & 67.2 {\scriptsize \textcolor{ForestGreen}{+09.0}} & \textbf{66.2} {\scriptsize \textcolor{ForestGreen}{+14.7}} \\
  \textcolor{RoyalBlue!80!black}{AbusEval} & 64.5 & 56.2 & 75.9 & 52.7 & 77.7 {\scriptsize \textcolor{ForestGreen}{+01.8}} & 51.5 {\scriptsize \textcolor{BrickRed}{-01.2}} & 77.7 {\scriptsize \textcolor{ForestGreen}{+01.8}} & 52.2 {\scriptsize \textcolor{BrickRed}{-00.5}} & 80.3 {\scriptsize \textcolor{ForestGreen}{+04.4}} & \textbf{66.1} {\scriptsize \textcolor{ForestGreen}{+13.4}} & 78.7 {\scriptsize \textcolor{ForestGreen}{+02.8}} & 61.7 {\scriptsize \textcolor{ForestGreen}{+09.0}} \\
  \textcolor{RoyalBlue!80!black}{AHSD} & 44.0 & 40.8 & 41.5 & 40.5 & 44.4 {\scriptsize \textcolor{ForestGreen}{+02.9}} & 43.6 {\scriptsize \textcolor{ForestGreen}{+03.1}} & 45.5 {\scriptsize \textcolor{ForestGreen}{+04.0}} & 44.5 {\scriptsize \textcolor{ForestGreen}{+04.0}} & 51.7 {\scriptsize \textcolor{ForestGreen}{+10.2}} & 49.7 {\scriptsize \textcolor{ForestGreen}{+09.2}} & 73.6 {\scriptsize \textcolor{ForestGreen}{+32.1}} & \textbf{66.3} {\scriptsize \textcolor{ForestGreen}{+25.8}} \\
  \midrule
  \textcolor{SeaGreen!80!black}{\textbf{GermEval21}} & 46.8 & 46.2 & 41.5 & 39.3 & 55.5 {\scriptsize \textcolor{ForestGreen}{+14.0}} & 55.4 {\scriptsize \textcolor{ForestGreen}{+16.1}} & 54.9 {\scriptsize \textcolor{ForestGreen}{+13.4}} & 54.8 {\scriptsize \textcolor{ForestGreen}{+15.5}} & 57.5 {\scriptsize \textcolor{ForestGreen}{+16.0}} & \textbf{57.3} {\scriptsize \textcolor{ForestGreen}{+18.0}} & 53.9 {\scriptsize \textcolor{ForestGreen}{+12.4}} & 53.9 {\scriptsize \textcolor{ForestGreen}{+14.6}} \\
  \textcolor{SeaGreen!80!black}{\textbf{GermEval19}} & 42.8 & 40.8 & 44.8 & 43.5 & 55.4 {\scriptsize \textcolor{ForestGreen}{+10.6}} & 55.4 {\scriptsize \textcolor{ForestGreen}{+11.9}} & 53.3 {\scriptsize \textcolor{ForestGreen}{+08.5}} & 53.1 {\scriptsize \textcolor{ForestGreen}{+09.6}} & 67.9 {\scriptsize \textcolor{ForestGreen}{+23.1}} & \textbf{66.8} {\scriptsize \textcolor{ForestGreen}{+23.3}} & 59.2 {\scriptsize \textcolor{ForestGreen}{+14.4}} & 59.1 {\scriptsize \textcolor{ForestGreen}{+15.6}} \\
  \textcolor{SeaGreen!80!black}{GermEval18} & 42.5 & 40.0 & 44.7 & 42.8 & 55.0 {\scriptsize \textcolor{ForestGreen}{+10.3}} & 54.9 {\scriptsize \textcolor{ForestGreen}{+12.1}} & 53.5 {\scriptsize \textcolor{ForestGreen}{+08.8}} & 53.3 {\scriptsize \textcolor{ForestGreen}{+10.5}} & 68.6 {\scriptsize \textcolor{ForestGreen}{+23.9}} & \textbf{67.2} {\scriptsize \textcolor{ForestGreen}{+24.4}} & 58.8 {\scriptsize \textcolor{ForestGreen}{+14.1}} & 58.7 {\scriptsize \textcolor{ForestGreen}{+15.9}} \\
  \textcolor{SeaGreen!80!black}{HASOC} & 40.5 & 40.2 & 56.5 & 54.1 & 56.7 {\scriptsize \textcolor{ForestGreen}{+00.2}} & 55.9 {\scriptsize \textcolor{ForestGreen}{+01.8}} & 53.6 {\scriptsize \textcolor{BrickRed}{-02.9}} & 53.2 {\scriptsize \textcolor{BrickRed}{-00.9}} & 69.8 {\scriptsize \textcolor{ForestGreen}{+13.3}} & \textbf{65.9} {\scriptsize \textcolor{ForestGreen}{+11.8}} & 61.6 {\scriptsize \textcolor{ForestGreen}{+05.1}} & 59.8 {\scriptsize \textcolor{ForestGreen}{+05.7}} \\
  \textcolor{SeaGreen!80!black}{Gahd} & 54.9 & 54.7 & 45.1 & 37.8 & 56.9 {\scriptsize \textcolor{ForestGreen}{+11.8}} & 55.3 {\scriptsize \textcolor{ForestGreen}{+17.5}} & 57.3 {\scriptsize \textcolor{ForestGreen}{+12.2}} & 55.7 {\scriptsize \textcolor{ForestGreen}{+17.9}} & 61.4 {\scriptsize \textcolor{ForestGreen}{+16.3}} & \textbf{61.3} {\scriptsize \textcolor{ForestGreen}{+23.5}} & 57.9 {\scriptsize \textcolor{ForestGreen}{+12.8}} & 56.8 {\scriptsize \textcolor{ForestGreen}{+19.0}} \\
  \midrule
  \textcolor{BurntOrange!90!black}{\textbf{ViHSD}} & 50.8 & 47.1 & 75.0 & 59.3 & 65.9 {\scriptsize \textcolor{BrickRed}{-09.1}} & 58.5 {\scriptsize \textcolor{BrickRed}{-00.8}} & 64.5 {\scriptsize \textcolor{BrickRed}{-10.5}} & 57.6 {\scriptsize \textcolor{BrickRed}{-01.7}} & 64.6 {\scriptsize \textcolor{BrickRed}{-10.4}} & 58.3 {\scriptsize \textcolor{BrickRed}{-01.0}} & 76.3 {\scriptsize \textcolor{ForestGreen}{+01.3}} & \textbf{66.1} {\scriptsize \textcolor{ForestGreen}{+06.8}} \\
  \midrule
  \textcolor{Orchid!80!black}{Haternet} & 35.3 & 34.9 & 44.2 & 44.0 & 38.6 {\scriptsize \textcolor{BrickRed}{-05.6}} & 38.5 {\scriptsize \textcolor{BrickRed}{-05.5}} & 36.3 {\scriptsize \textcolor{BrickRed}{-07.9}} & 35.9 {\scriptsize \textcolor{BrickRed}{-08.1}} & 52.3 {\scriptsize \textcolor{ForestGreen}{+08.1}} & 49.8 {\scriptsize \textcolor{ForestGreen}{+05.8}} & 50.9 {\scriptsize \textcolor{ForestGreen}{+06.7}} & \textbf{50.5} {\scriptsize \textcolor{ForestGreen}{+06.5}} \\
  \textcolor{Orchid!80!black}{HateEval-spa} & 42.6 & 38.8 & 53.7 & 53.7 & 46.3 {\scriptsize \textcolor{BrickRed}{-07.4}} & 42.9 {\scriptsize \textcolor{BrickRed}{-10.8}} & 47.2 {\scriptsize \textcolor{BrickRed}{-06.5}} & 43.3 {\scriptsize \textcolor{BrickRed}{-10.4}} & 49.9 {\scriptsize \textcolor{BrickRed}{-03.8}} & 49.8 {\scriptsize \textcolor{BrickRed}{-03.9}} & 54.4 {\scriptsize \textcolor{ForestGreen}{+00.7}} & \textbf{53.9} {\scriptsize \textcolor{ForestGreen}{+00.2}} \\
  \textcolor{Orchid!80!black}{Chileno} & 25.4 & 23.7 & 47.6 & 37.7 & 29.0 {\scriptsize \textcolor{BrickRed}{-18.6}} & 25.9 {\scriptsize \textcolor{BrickRed}{-11.8}} & 26.6 {\scriptsize \textcolor{BrickRed}{-21.0}} & 24.2 {\scriptsize \textcolor{BrickRed}{-13.5}} & 51.6 {\scriptsize \textcolor{ForestGreen}{+04.0}} & \textbf{39.0} {\scriptsize \textcolor{ForestGreen}{+01.3}} & 42.9 {\scriptsize \textcolor{BrickRed}{-04.7}} & 35.6 {\scriptsize \textcolor{BrickRed}{-02.1}} \\
  \midrule
  \textbf{Avg} & 47.5 & 47.4 & 55.6 & 54.8 & 58.1 {\scriptsize \textcolor{ForestGreen}{+02.5}} & 56.9 {\scriptsize \textcolor{ForestGreen}{+02.1}} & 57.4 {\scriptsize \textcolor{ForestGreen}{+01.8}} & 56.5 {\scriptsize \textcolor{ForestGreen}{+01.7}} & 64.6 {\scriptsize \textcolor{ForestGreen}{+09.0}} & 63.1 {\scriptsize \textcolor{ForestGreen}{+08.3}} & 65.7 {\scriptsize \textcolor{ForestGreen}{+10.1}} & \textbf{65.4} {\scriptsize \textcolor{ForestGreen}{+10.6}} \\
  \bottomrule
    \end{tabular}
    }
    \end{subtable}
    \vspace{0.5em}
    \begin{subtable}[t]{\linewidth}
    \centering
    \caption*{\texttt{Qwen2.5-14B}}
    \resizebox{\linewidth}{!}{
    \begin{tabular}{|lcc|cc|cc|cc|cc|cc|} 
        \toprule
        \textbf{Dataset} & \multicolumn{2}{c|}{\textbf{Base}} &
        \multicolumn{2}{c|}{\textbf{Human}} &
        \multicolumn{2}{c|}{\textbf{Mean}} &
        \multicolumn{2}{c|}{\textbf{Voting}} &
        \multicolumn{2}{c|}{\textbf{LGB}} &
        \multicolumn{2}{c|}{\textbf{Human+LGB}} \\
        \cmidrule(lr){2-3} \cmidrule(lr){4-5} \cmidrule(lr){6-7} \cmidrule(lr){8-9} \cmidrule(lr){10-11} \cmidrule(lr){12-13}
        &  Acc & F1 &   Acc & F1 &  Acc & F1 &  Acc & F1  & Acc & F1  & Acc & F1 \\
        \midrule
        \textcolor{RoyalBlue!80!black}{\textbf{HateXplain}} & 66.3 & 54.6 & 73.9 & 70.0 & 74.9 {\scriptsize \textcolor{ForestGreen}{+01.0}} & 72.2 {\scriptsize \textcolor{ForestGreen}{+02.2}} & 73.8 {\scriptsize \textcolor{BrickRed}{-00.1}} & 72.4 {\scriptsize \textcolor{ForestGreen}{+02.4}} & 73.2 {\scriptsize \textcolor{BrickRed}{-00.7}} & 69.5 {\scriptsize \textcolor{BrickRed}{-00.5}} & 75.6 {\scriptsize \textcolor{ForestGreen}{+01.7}} & \textbf{73.1} {\scriptsize \textcolor{ForestGreen}{+03.1}} \\
        \textcolor{RoyalBlue!80!black}{\textbf{Sexism}} & 75.0 & 65.0 & 86.1 & \textbf{73.8} & 85.8 {\scriptsize \textcolor{BrickRed}{-00.3}} & 64.9 {\scriptsize \textcolor{BrickRed}{-08.9}} & 85.7 {\scriptsize \textcolor{BrickRed}{-00.4}} & 62.9 {\scriptsize \textcolor{BrickRed}{-10.9}} & 86.5 {\scriptsize \textcolor{ForestGreen}{+00.4}} & 69.3 {\scriptsize \textcolor{BrickRed}{-04.5}} & 88.0 {\scriptsize \textcolor{ForestGreen}{+01.9}} & 71.1 {\scriptsize \textcolor{BrickRed}{-02.7}} \\
        \textcolor{RoyalBlue!80!black}{\textbf{Covid}} & 39.9 & 39.8 & 66.5 & 62.3 & 65.9 {\scriptsize \textcolor{BrickRed}{-00.6}} & 62.4 {\scriptsize \textcolor{ForestGreen}{+00.1}} & 72.6 {\scriptsize \textcolor{ForestGreen}{+06.1}} & 67.6 {\scriptsize \textcolor{ForestGreen}{+05.3}} & 69.7 {\scriptsize \textcolor{ForestGreen}{+03.2}} & 65.8 {\scriptsize \textcolor{ForestGreen}{+03.5}} & 72.6 {\scriptsize \textcolor{ForestGreen}{+06.1}} & \textbf{67.9} {\scriptsize \textcolor{ForestGreen}{+05.6}} \\
        \textcolor{RoyalBlue!80!black}{\textbf{US\_election}} & 49.6 & 46.0 & 86.8 & 66.1 & 83.1 {\scriptsize \textcolor{BrickRed}{-03.7}} & \textbf{67.1} {\scriptsize \textcolor{ForestGreen}{+01.0}} & 85.3 {\scriptsize \textcolor{BrickRed}{-01.5}} & 62.3 {\scriptsize \textcolor{BrickRed}{-03.8}} & 85.4 {\scriptsize \textcolor{BrickRed}{-01.4}} & 66.8 {\scriptsize \textcolor{ForestGreen}{+00.7}} & 86.3 {\scriptsize \textcolor{BrickRed}{-00.5}} & 63.3 {\scriptsize \textcolor{BrickRed}{-02.8}} \\
        \textcolor{RoyalBlue!80!black}{HateEval-eng} & 63.0 & 61.7 & 67.2 & 66.9 & 69.7 {\scriptsize \textcolor{ForestGreen}{+02.5}} & 69.5 {\scriptsize \textcolor{ForestGreen}{+02.6}} & 69.0 {\scriptsize \textcolor{ForestGreen}{+01.8}} & 68.3 {\scriptsize \textcolor{ForestGreen}{+01.4}} & 70.0 {\scriptsize \textcolor{ForestGreen}{+02.8}} & \textbf{69.8} {\scriptsize \textcolor{ForestGreen}{+02.9}} & 70.0 {\scriptsize \textcolor{ForestGreen}{+02.8}} & 69.4 {\scriptsize \textcolor{ForestGreen}{+02.5}} \\
        \textcolor{RoyalBlue!80!black}{AbusEval} & 66.9 & 63.4 & 82.1 & \textbf{68.7} & 81.5 {\scriptsize \textcolor{BrickRed}{-00.6}} & 68.2 {\scriptsize \textcolor{BrickRed}{-00.5}} & 81.4 {\scriptsize \textcolor{BrickRed}{-00.7}} & 67.2 {\scriptsize \textcolor{BrickRed}{-01.5}} & 80.5 {\scriptsize \textcolor{BrickRed}{-01.6}} & 67.1 {\scriptsize \textcolor{BrickRed}{-01.6}} & 82.1 {\scriptsize \textcolor{ForestGreen}{+00.0}} & 66.2 {\scriptsize \textcolor{BrickRed}{-02.5}} \\
        \textcolor{RoyalBlue!80!black}{AHSD} & 91.9 & \textbf{84.9} & 81.9 & 74.9 & 81.5 {\scriptsize \textcolor{BrickRed}{-00.4}} & 74.8 {\scriptsize \textcolor{BrickRed}{-00.1}} & 63.8 {\scriptsize \textcolor{BrickRed}{-18.1}} & 59.6 {\scriptsize \textcolor{BrickRed}{-15.3}} & 72.8 {\scriptsize \textcolor{BrickRed}{-09.1}} & 67.1 {\scriptsize \textcolor{BrickRed}{-07.8}} & 79.5 {\scriptsize \textcolor{BrickRed}{-02.4}} & 72.9 {\scriptsize \textcolor{BrickRed}{-02.0}} \\
        \midrule
        \textcolor{SeaGreen!80!black}{\textbf{GermEval21}} & 64.4 & \textbf{63.8} & 68.9 & \textbf{63.8} & 68.1 {\scriptsize \textcolor{BrickRed}{-00.8}} & 58.5 {\scriptsize \textcolor{BrickRed}{-05.3}} & 66.2 {\scriptsize \textcolor{BrickRed}{-02.7}} & 51.3 {\scriptsize \textcolor{BrickRed}{-12.5}} & 67.3 {\scriptsize \textcolor{BrickRed}{-01.6}} & 58.3 {\scriptsize \textcolor{BrickRed}{-05.5}} & 68.6 {\scriptsize \textcolor{BrickRed}{-00.3}} & 59.9 {\scriptsize \textcolor{BrickRed}{-03.9}} \\
        \textcolor{SeaGreen!80!black}{\textbf{GermEval19}} & 72.2 & 71.8 & 81.3 & 78.9 & 81.9 {\scriptsize \textcolor{ForestGreen}{+00.6}} & 79.0 {\scriptsize \textcolor{ForestGreen}{+00.1}} & 79.5 {\scriptsize \textcolor{BrickRed}{-01.8}} & 74.0 {\scriptsize \textcolor{BrickRed}{-04.9}} & 81.2 {\scriptsize \textcolor{BrickRed}{-00.1}} & 78.3 {\scriptsize \textcolor{BrickRed}{-00.6}} & 82.2 {\scriptsize \textcolor{ForestGreen}{+00.9}} & \textbf{79.1} {\scriptsize \textcolor{ForestGreen}{+00.2}} \\
        \textcolor{SeaGreen!80!black}{GermEval18} & 75.8 & 75.3 & 82.8 & \textbf{80.9} & 82.6 {\scriptsize \textcolor{BrickRed}{-00.2}} & 79.5 {\scriptsize \textcolor{BrickRed}{-01.4}} & 80.1 {\scriptsize \textcolor{BrickRed}{-02.7}} & 75.0 {\scriptsize \textcolor{BrickRed}{-05.9}} & 82.6 {\scriptsize \textcolor{BrickRed}{-00.2}} & 79.9 {\scriptsize \textcolor{BrickRed}{-01.0}} & 83.2 {\scriptsize \textcolor{ForestGreen}{+00.4}} & 80.5 {\scriptsize \textcolor{BrickRed}{-00.4}} \\
        \textcolor{SeaGreen!80!black}{HASOC} & 67.7 & 66.6 & 81.7 & 78.1 & 82.7 {\scriptsize \textcolor{ForestGreen}{+01.0}} & \textbf{78.7} {\scriptsize \textcolor{ForestGreen}{+00.6}} & 83.1 {\scriptsize \textcolor{ForestGreen}{+01.4}} & 76.8 {\scriptsize \textcolor{BrickRed}{-01.3}} & 81.6 {\scriptsize \textcolor{BrickRed}{-00.1}} & 75.5 {\scriptsize \textcolor{BrickRed}{-02.6}} & 83.1 {\scriptsize \textcolor{ForestGreen}{+01.4}} & 78.3 {\scriptsize \textcolor{ForestGreen}{+00.2}} \\
        \textcolor{SeaGreen!80!black}{Gahd} & 73.5 & 73.5 & 77.7 & \textbf{77.7} & 75.8 {\scriptsize \textcolor{BrickRed}{-01.9}} & 75.0 {\scriptsize \textcolor{BrickRed}{-02.7}} & 71.6 {\scriptsize \textcolor{BrickRed}{-06.1}} & 69.3 {\scriptsize \textcolor{BrickRed}{-08.4}} & 75.5 {\scriptsize \textcolor{BrickRed}{-02.2}} & 75.0 {\scriptsize \textcolor{BrickRed}{-02.7}} & 77.3 {\scriptsize \textcolor{BrickRed}{-00.4}} & 77.0 {\scriptsize \textcolor{BrickRed}{-00.7}} \\
        \midrule
        \textcolor{BurntOrange!90!black}{\textbf{ViHSD}} & 78.7 & 71.9 & 85.0 & 76.0 & 87.1 {\scriptsize \textcolor{ForestGreen}{+02.1}} & 76.0 {\scriptsize \textcolor{ForestGreen}{+00.0}} & 86.4 {\scriptsize \textcolor{ForestGreen}{+01.4}} & 70.9 {\scriptsize \textcolor{BrickRed}{-05.1}} & 87.3 {\scriptsize \textcolor{ForestGreen}{+02.3}} & 75.9 {\scriptsize \textcolor{BrickRed}{-00.1}} & 86.9 {\scriptsize \textcolor{ForestGreen}{+01.9}} & \textbf{77.5} {\scriptsize \textcolor{ForestGreen}{+01.5}} \\
        \midrule
        \textcolor{Orchid!80!black}{Haternet} & 46.4 & 46.3 & 69.5 & 67.3 & 77.4 {\scriptsize \textcolor{ForestGreen}{+07.9}} & 74.0 {\scriptsize \textcolor{ForestGreen}{+06.7}} & 81.2 {\scriptsize \textcolor{ForestGreen}{+11.7}} & \textbf{76.0} {\scriptsize \textcolor{ForestGreen}{+08.7}} & 75.0 {\scriptsize \textcolor{ForestGreen}{+05.5}} & 72.2 {\scriptsize \textcolor{ForestGreen}{+04.9}} & 77.0 {\scriptsize \textcolor{ForestGreen}{+07.5}} & 73.5 {\scriptsize \textcolor{ForestGreen}{+06.2}} \\
        \textcolor{Orchid!80!black}{HateEval-spa} & 57.9 & 54.6 & 65.2 & 64.6 & 65.2 {\scriptsize \textcolor{ForestGreen}{+00.0}} & 65.1 {\scriptsize \textcolor{ForestGreen}{+00.5}} & 67.8 {\scriptsize \textcolor{ForestGreen}{+02.6}} & \textbf{67.8} {\scriptsize \textcolor{ForestGreen}{+03.2}} & 67.2 {\scriptsize \textcolor{ForestGreen}{+02.0}} & 67.0 {\scriptsize \textcolor{ForestGreen}{+02.4}} & 67.5 {\scriptsize \textcolor{ForestGreen}{+02.3}} & 67.4 {\scriptsize \textcolor{ForestGreen}{+02.8}} \\
        \textcolor{Orchid!80!black}{Chileno} & 32.6 & 29.2 & 60.1 & 44.1 & 60.3 {\scriptsize \textcolor{ForestGreen}{+00.2}} & 44.3 {\scriptsize \textcolor{ForestGreen}{+00.2}} & 74.1 {\scriptsize \textcolor{ForestGreen}{+14.0}} & \textbf{49.6} {\scriptsize \textcolor{ForestGreen}{+05.5}} & 63.4 {\scriptsize \textcolor{ForestGreen}{+03.3}} & 46.8 {\scriptsize \textcolor{ForestGreen}{+02.7}} & 62.1 {\scriptsize \textcolor{ForestGreen}{+02.0}} & 44.8 {\scriptsize \textcolor{ForestGreen}{+00.7}} \\
        \midrule
        \textbf{Avg} & 67.8 & 67.8 & 77.2 & 76.1 & 77.5 {\scriptsize \textcolor{ForestGreen}{+00.3}} & 75.9 {\scriptsize \textcolor{BrickRed}{-00.2}} & 76.1 {\scriptsize \textcolor{BrickRed}{-01.1}} & 72.8 {\scriptsize \textcolor{BrickRed}{-03.3}} & 76.7 {\scriptsize \textcolor{BrickRed}{-00.5}} & 75.1 {\scriptsize \textcolor{BrickRed}{-01.0}} & 78.3 {\scriptsize \textcolor{ForestGreen}{+01.1}} & \textbf{76.7} {\scriptsize \textcolor{ForestGreen}{+00.6}} \\
        \bottomrule
    \end{tabular}
    }
    \end{subtable}
    \caption{Full per-dataset accuracy and macro-F1~(\%) of \texttt{Llama3.2-1B} (top) and \texttt{Qwen2.5-14B} (bottom) under six training conditions.
    \texttt{Base} = zero-shot; \texttt{Human} = 7-Set human labels; \texttt{Mean}/\texttt{Voting}/\texttt{LGB} = synthetic-only; \texttt{Human+LGB} = combined.
    Signed values ($\pm$) are relative to \texttt{Human}.  Bold~= best F1 per row.
    The classification threshold is the mean predicted probability of the \textit{Hate} class, not the default~0.5.}
    \label{tab:1B-full}
\end{table*}

\end{document}